# Efficient inference in persistent Dynamic Bayesian Networks

Tomáš Šingliar
Department of Computer Science
University of Pittsburgh
Pittsburgh, PA 15213

Denver H. Dash
Intel Research and
Department of Biomedical Informatics
University of Pittsburgh
Pittsburgh, PA 15213

## Abstract

Numerous temporal inference tasks such as fault monitoring and anomaly detection exhibit a *persistence* property: for example, if something breaks, it stays broken until an intervention. When modeled as a Dynamic Bayesian Network, persistence adds dependencies between adjacent time slices, often making exact inference over time intractable using standard inference algorithms. However, we show that persistence implies a regular structure that can be exploited for efficient inference. We present three successively more general classes of models: *persistent causal chains* (PCCs), *persistent causal trees* (PCTs) and *persistent polytrees* (PPTs), and the corresponding exact inference algorithms that exploit persistence. We show that analytic asymptotic bounds for our algorithms compare favorably to junction tree inference; and we demonstrate empirically that we can perform exact smoothing on the order of 100 times faster than the approximate Boyen-Koller method on randomly generated instances of persistent tree models. We also show how to handle non-persistent variables and how persistence can be exploited effectively for approximate filtering.

## 1 Introduction

Persistence is a common trait of many real-world systems. It is used to model permanent changes in state, such as when components of a system that have broken until someone intervenes to fix them. Especially interesting and useful are diagnostic models where misalignments and other process drifts may cause a cascade of other failures, all of which may also persist until the root cause is fixed. Even when such changes are not truly permanent, they are often reversed slowly relative to the time scale of the model, and persistence can be a good approximation in such systems. For instance, vehicular accidents cause obstructions on the road that last much longer than the required detection time and are thus persistent for the purpose of detection [20]. Another example is outbreak detection [4], where an infected population stays infected much longer than the desired detection time. There are many other examples of persistence and approximate persistence.

Dynamic Bayesian Networks (DBNs) [5] are a general formalism for modeling temporal systems under uncertainty. Many standard time-series methods are special cases of DBNs, including Hidden Markov Models [18] and Kalman filters [7]. Discrete DBNs in particular are a very popular formalism, but usually suffer from intractability [1] when dense inter-temporal dependencies are present among hidden state variables, leading many to search for approximation algorithms [1, 13, 15, 14]. Unfortunately, modeling persistence with DBNs requires the introduction of many inter-temporal arcs, often making exact inference intractable with standard inference algorithms.

In this paper, we define Persistent Causal DBNs (PC-DBNs), a particular class of DBN models capable of modeling many real-world systems that involve long chains of causal influence coupled with persistence of causal effects. We show that a linear time algorithm exists for inference (smoothing) in linear chain and tree-based PC-DBNs. We then generalize our results to polytree causal networks, where the algorithm remains exact, and to general networks, where it inherits its properties of loopy belief propagation [21]. Our method relies on a transformation of the original prototype network, allowing smoothing to be done efficiently; however, this method does not readily deal with the incremental filtering problem. Nonetheless, we show empirically that, if evidence is observed at every time slice, approximate filtering can be accomplished with fixed window smoothing, producing lower error than approximate Boyen-Koller (BK) filtering [1] using a fraction of the computation time.

The algorithm that we present exploits a particular type of determinism that is given by the persistence relation. There has been other work that seeks to directly or indirectly exploit general deterministic structure in Bayesian networks using compilation approaches [2], a generalized version belief propagation [10], and variable elimination with algebraic decision diagrams [3, 19]. These more general methods have not been tailored to the important special cases of DBNs and persistency. To our knowledge, this is the first work to investigate persistency in DBNs.

The paper is organized as follows: In Section 2 we introduce the changepoint transformation. Section 3 introduces persistent causal chain DBNs and the corresponding inference algorithm, which retains all the essential properties of later models. Then, Section 4 will discuss the steps leading to a fully general algorithm. Experimental results are presented in Section 5, followed by conclusions.

## 2 Notation and changepoints

Consider a Bayesian network (BN) with $N$ binary variables $X_i$; we will refer to this network as the *prototype*. The corresponding Dynamic BN with $M$ slices is created by replicating the prototype $M$ times and connecting some of the variables to their copies in the next slice. In our notation, upper indices range over time slices of the DBN; lower indices range over variables in each time slice. Colon notation is used to denote sets and sequences. Thus, for instance, $X_4^{1:M}$ denotes the entire temporal sequence of values of $X_4$ from time 1 to time $M$. Variables without an upper index will refer to their respective counterparts in the prototype. We say that a variable $X_k$ is *persistent* if

$$P(X_k^t = 1 | X_k^{t-1}, U^t) = \begin{cases} P(X_k|U) & \text{if } X_k^{t-1} = 0 \\ 1 & \text{if } X_k^{t-1} = 1 \end{cases}, \quad (1)$$

where $U = Pa(X_k)$ refers to the parents of $X_k$ in the prototype. In other words, 1 is an absorbing state. Sometimes [12] a variable is called persistent if it has an arc to the next-slice copy of itself. Our definition of *persistence* is strictly stronger, but no confusion should arise in this paper.

There are $2^M$ temporal sequences of values of a binary variable $X_k$. If the variable is persistent, the number of configurations is reduced to $M+1$. Information about $X_k^{1:M}$ can be summarized by looking at the time when $X$ changed from 0 to 1 (we sometimes refer to the 0 state as the *off* state and 1 as the *on* state). Thus, inference in the persistent DBN with binary variables is equivalent to inference in a network whose topology closely resembles that of the prototype and whose variables are $M+1$-ary discrete *changepoint* variables, with correspondingly defined conditional probability distributions (CPDs), as shown in Figure 1b. The models in Figure 1a and 1b are identical; one can go back and forth between them by recognizing that

$$(\tilde{X} = j) \Leftrightarrow (X^j = 0) \wedge (X^{j+1} = 1) \text{ and}$$
$$(X^j = 0) \Leftrightarrow (\tilde{X} > j).$$

If the prototype is a tree, belief propagation in the transformed network yields an algorithm whose complexity is $O(M^2 N)$. The quadratic part of the computation comes from summing over the $M+1$ values of the single parent for each of the $M+1$ values of the child. Similarly, if the prototype is a polytree, complexity will be proportional to $M^{U_{max}+1}$, where $U_{max}$ is the largest in-degree in the network. This transformation by itself, when all hidden state variables are persistent, allows us to perform smoothing much more efficiently than by operating on the original DBN. There is, however, additional structure in the CPDs that allows us to do better by a factor of $M$, and we can also adapt our algorithm to deal with the case when some hidden variables are not persistent.

## 3 PCC-DBN inference

To simplify the exposition, let us now focus on a specific prototype, a persistent causal chain DBN (PCC-DBN). This is a chain with $Pa(X_i) = \{X_{i-1}\}, i = 1, ..., N$ and $Pa(O) = X_N$ (thus it has N+1 nodes). Let us further assume that the leaves are non-persistent and observed, while the causes ($X$ nodes) are all persistent and hidden. The network is shown in Figure 1a and its transformed version in Figure 1b.

Consider the problem of computing $P(O)$. This is in general one of the most difficult inference problems, requiring one to integrate out all hidden state variables, and is implicit in most inference queries:

$$P(O^{1:M}) = \sum_{X_{1:N}^{1:M}} P(O^{1:M} | X_{1:N}^{1:M}) \cdot P(X_{1:N}^{1:M}) \quad (2)$$

Let $\{j_k : 0 \leq j_k \leq M\}$ index the sequence of $X_k^{1:M}$ in which variable $X_k^{j_k}$ is the last (highest-time) variable to be in the *off* state, unless $j_k = 0$ in which case it indexes the sequence in which all $X_k$ are in the *on* state. As an example, if $M = 3$, then $j_k = \{0, 1, 2, 3\}$ indexes the states $X_k^{1:M} = \{111, 011, 001, 000\}$, respectively, for all $k$. All configurations not indexed by $j_i$ have zero probability due to the persistence assumption. To simplify notation, we use $j_k$ to denote the event that $X_k^{1:M}$ is the sequence indexed by $j_k$. We also say that $X_k$ *fired* at $j_k$. We can decompose Equa-

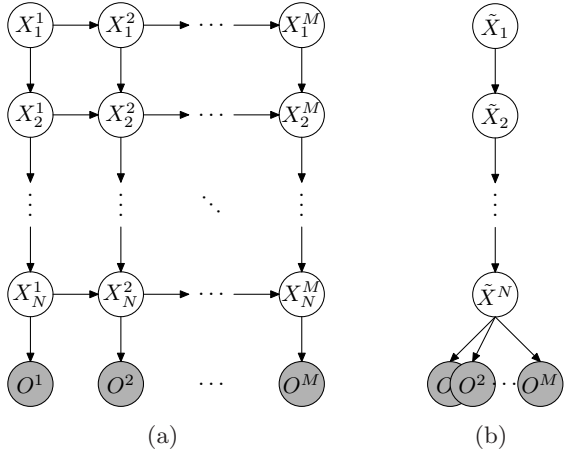

Figure 1: **(a)** A PCC-DBN network with $N+1$ nodes per slice; **(b)** the transformed network. We sometimes refer to $X_2^1$ as the temporal parent of $X_2^2$ and to $X_1^2$ as its causal parent.

tion 2 according to the network structure as follows:

$$P(O^{1:M}) = \sum_{j_1=0}^{M} P(j_1) \sum_{j_2=0}^{M} P(j_2 \mid j_1) \ldots$$
$$\ldots \sum_{j_N=0}^{M} P(j_N \mid j_{N-1}) \cdot P(O^{1:M} \mid j_N) \quad (3)$$

Denote by $P_k$ the probability that variable $X_k$ will fire for the first time given that its causal parent *has* fired, and by $\widehat{P}_k$ the probability that $X_k$ will fire for the first time given that its causal parent has not fired:

$$P_k \equiv P(X_k^j = 1 \mid X_k^{j-1} = 0, X_{k-1}^j = 1)$$
$$\widehat{P}_k \equiv P(X_k^j = 1 \mid X_k^{j-1} = 0, X_{k-1}^j = 0).$$

Let $P_{\bar{k}}$ and $\widehat{P}_{\bar{k}}$ denote the complements $1 - P_k$ and $1 - \widehat{P}_k$, respectively. We can define $\Sigma_k^L$ recursively to denote the partial sum over $j_k$ from Equation 3, conditioned on $j_{k-1} = L$:

$$\Sigma_k^L \equiv \sum_{j_k} P(j_k \mid j_{k-1} = L) \cdot \Sigma_{k+1}^{j_k} \quad (4)$$

with boundary condition $\Sigma_{N+1}^L \equiv P(O^{1:M} \mid j_N = L)$. Using this notation, Equation 3 can be rewritten as:

$$P(O^{1:M}) = \sum_{j_1=0}^{M} P(j_1) \cdot \Sigma_2^{j_1} \quad (5)$$

Now we now need to show that one can calculate the entire set $\Sigma_{2:N}^{0:M}$ in time $O(MN)$. Each $\Sigma_k^L$ can be written as follows:

$$\Sigma_k^L = \bar{\sigma}_k^L + \sigma_k^L + \hat{\sigma}_k^L, \quad (6)$$

where $\bar{\sigma}_k^L$ contains all the terms in the sum such that $X_k$ first fires when $X_{k-1}$ has not fired:

$$\bar{\sigma}_k^L = \sum_{j_k < L} \widehat{P}_{\bar{k}}^{j_k} \widehat{P}_k \cdot \Sigma_{k+1}^{j_k}. \quad (7)$$

$\sigma_k^L$ contains all the terms in which $X_k$ first fires when $X_{k-1}$ has *also* fired:

$$\sigma_k^L = \sum_{L \leq j_k < M} \widehat{P}_{\bar{k}}^{L} P_{\bar{k}}^{j_k - L} P_k \cdot \Sigma_{k+1}^{j_k}, \quad (8)$$

and $\hat{\sigma}_k^L$ contains the final term in which $X_k$ never fires:

$$\hat{\sigma}_k^L = \widehat{P}_{\bar{k}}^{L} P_{\bar{k}}^{M-L} \cdot \Sigma_{k+1}^{M}. \quad (9)$$

In order to calculate Equation 5 in time $O(MN)$, we need to pre-compute $\bar{\sigma}_k^L$, $\sigma_k^L$ and $\hat{\sigma}_k^L$ for all values of $L$ in $O(M)$ for each variable $X_k$.

### 3.1 Upward Recursion Relations

As a boundary condition for the recursion, assume we have calculated $\Sigma_{N+1}^k$ for all $0 \leq k \leq M$. We show how to do this in time $O(M)$ in Section 3.2. Also, this algorithm requires the pre-calculation and caching of $\widehat{P}_{\bar{k}}^i$ for $0 \leq k \leq N$ and $0 \leq i \leq M$, which can be done recursively in $O(MN)$ time and space.

Inspecting Equation 7 more closely, it should be easy to see that one can calculate $\bar{\sigma}_N^i$ for $0 \leq i \leq M$ in $O(M)$ time using the following recursion:

$$\bar{\sigma}_l^{i+1} = \bar{\sigma}_l^i + \widehat{P}_{\bar{l}}^i \cdot \widehat{P}_l \cdot \Sigma_{k+1}^i, \quad (10)$$

with boundary condition $\bar{\sigma}_l^0 = 0$ for all $l$. One can also calculate $\sigma_k^i$ for $0 \leq i \leq M$ with the recursion:

$$\sigma_k^{i-1} = \frac{\sigma_k^i}{\widehat{P}_{\bar{k}}} P_{\bar{k}} + \widehat{P}_{\bar{k}}^{i-1} \cdot P_k \cdot \Sigma_{k+1}^{i-1}, \quad (11)$$

with boundary condition $\sigma_k^M = 0$ for all $l$. Finally, one can calculate $\hat{\sigma}_N^i$ for $0 \leq i \leq M$ with the recursion:

$$\hat{\sigma}_k^{i-1} = \frac{\hat{\sigma}_k^i}{\widehat{P}_{\bar{k}}} P_{\bar{k}} \quad (12)$$

with boundary condition $\hat{\sigma}_k^M = \widehat{P}_{\bar{k}}^M \cdot \Sigma_{k+1}^M$ for all $l$.

Once $\bar{\sigma}_N^i$, $\sigma_N^i$ and $\hat{\sigma}_N^i$ are calculated, one can calculate all $\Sigma_N^i$ for $0 \leq i \leq M$ in $O(M)$ time using Equations 10, 11, 12 and 6. After $\Sigma_N^{0:M}$ is calculated, we can use Equation 4 to obtain $\Sigma_{N-1}^{0:M}$ in time $O(M)$, and repeat $N$ times to get all values of $\Sigma_{1:N}^{0:M}$. Thus the entire calculation takes $O(MN)$ time.

## 3.2 Computing $\Sigma^i_{N+1}$

To finalize the proof, we have to show how to calculate $\Sigma^i_{N+1}$ (the probability of the observations for a given configuration $i$ of $X^{1:M}_N$) for *all* $0 \leq i \leq M$ in time $O(M)$. Recall that $\Sigma^i_{N+1} \equiv P(O^{1:M} \mid j_N = i)$. Since the parent of each $O^j$ is given, for each $i$, this calculation is simply the product of the observations:

$$P(O^{1:M} \mid j_N = i) = \prod_{k=1}^{M} P(O^k \mid X^k_N, j_N = i) \quad (13)$$

Using our existing notation, we define

$$\phi^\ell_{N+1} = P(O^k \mid X^k_N = 1), \quad (14)$$
$$\bar{\phi}^\ell_{N+1} = P(O^k \mid X^k_N = 0), \quad (15)$$

$\Sigma^i_{N+1}$ can be calculated for all $0 \leq i \leq M$ in time $O(M)$ via the recursion relation:

$$\Sigma^0_{N+1} = \prod_{\ell=1}^{M} \phi^\ell_{N+1} \quad \text{and} \quad \Sigma^{\ell+1}_{N+1} = \Sigma^\ell_{N+1} \cdot \frac{\bar{\phi}^\ell_{N+1}}{\phi^\ell_{N+1}}. \quad (16)$$

Note that this formulation puts no distributional assumption on $P(O|X_N)$. The leaves can be distributed as multinomials, Gaussians etc, as is often done with Hidden Markov models [18] when they are put to their many uses.

## 3.3 Downward Recurrences

The above discussion completes the description of the "$\lambda$-pass" of PCC-DBN algorithm. Similar reasoning can be applied to obtain the "$\pi$-pass" recurrences that we now give without full derivation. Analogously to $\Sigma$, the semantics of $\Psi^j_k$ is $p(X_k = j|O^+_k)$, where $O^+_k$ is the subset of evidence reachable from $X_k$ through its parent[1]. $\Psi^j_k$ is again a sum of three components:

$$\Psi^j_k = \psi^j_k + \bar{\psi}^j_k + \hat{\psi}^j_k \quad (17)$$

$\bar{\psi}$ accounts for the terms where the parent has not yet changed:

$$\bar{\psi}^{\ell-1}_k = \bar{\psi}^\ell_k \cdot \frac{1}{P_{\bar{k}}} + \widehat{P}^{\ell-1}_{\bar{k}} \cdot \widehat{P}_k \cdot \Psi^\ell_{k-1} \quad (18)$$

with initialization $\bar{\psi}^M_k = 0$ for all $k$.

$\psi$ accounts for the terms where the parent has already changed:

$$\psi^{\ell+1}_k = \psi^\ell_k \cdot P_{\bar{k}} + \widehat{P}^{\ell+1}_{\bar{k}} \cdot P_k \cdot \Psi^{\ell+1}_{k-1} \quad (19)$$

---
[1] We only have evidence in the bottom layer in PCC-DBNs, but this will come handy in the next section.

with boundary condition $\psi^0_k = P_k \Psi^0_{k-1}$ for all $k$. Also, since $X_k$ eventually changes in this scenario, $\psi^M_k = 0$.

$\hat{\psi}$ accounts for the terms where the node never changes:

$$\hat{\psi}^M_k = \sum_{0 \leq i \leq M} \widehat{P}^i_{\bar{k}} \cdot P^{M-i}_{\bar{k}} \cdot \Psi^i_{k-1}. \quad (20)$$

Because the upper index refers to the changepoint of $X_k$, only $\hat{\psi}^M_k$ is non-zero. We can just compute this in $O(M)$ without the need for recurrences.

Initialization of the $\Psi$-recurrences happens at the root(s) of the network. For any root $r$, $\Psi^0_r = \widehat{P}_k$ and recurrently $\Psi^{i+1}_r = \Psi^i_r \cdot \widehat{P}_{\bar{r}}$. Finally, $\Psi^M_r = \Psi^{M-1}_r \widehat{P}_{\bar{r}}/\widehat{P}_k$.

## 3.4 PCC-DBN and belief propagation

We have just defined PCC-DBN, a version of belief propagation that first collects the evidence by passing the $\lambda$-messages towards the root of the chain and the proceeds to distribute information towards the leaves via the $\pi$-messages. After propagation is complete, we can obtain any posterior as

$$p(\tilde{X}_k|O) = \Sigma_{k+1} \cdot \Psi_k. \quad (21)$$

It is now useful to recall the types of potentials involved in Pearl's algorithm [8] and how they relate to the quantities above. For each node $X$, there are local potentials $\pi_X(x) \equiv_{def} p(X = x|e^+_X)$ and $\lambda_X(x) \equiv_{def} p(e^-_X|X = x)$, where $e^+_X$ and $e^-_X$ denote respectively the evidence reachable through parents and the evidence reachable from $X$ "downwards", $X$ included. There are two types of messages in Pearl's algorithm: $\pi_{X \to Y_i}$ sent by $X$ to its children and $\lambda_{X \to U_i}$ sent to its parents. A closer look at PCC-DBN reveals that each $\Sigma_k$ is identical to $\lambda_{X_k \to X_{k-1}}$ — the message from $X_k$ to its single parent $X_{k-1}$. The local potential $\lambda_{X_k}(j_k)$ is identical to $\Sigma_{k+1}$, because there are no children other than $X_{k+1}$ and evidence is only observed at the bottom of the chain. $\Psi_k$ corresponds directly to $\pi_{X_k}(j_k)$. This is why Equation 21 works.

## 3.5 Simple Generalizations and Causal Trees

While PCC-DBNs are useful for demonstrating the general ideas of handling the probability distributions arising from the changepoint transformation, they form a rather restricted class of networks, and the inference query that we performed was also restricted. Here we state succinctly a set of simple alterations which allow this algorithm to be relaxed in various ways:

**General evidence patterns** We can have observations anywhere in the network, in any time slice.

Casting the inference as belief propagation gives the answer to any probabilistic query as Equation 21 with one caveat: An observation such as $X_3^3 = 1$ does not tell us with certainty the position of the changepoint, but it just provides evidence that $j_3 < 3$, thus we cannot simply set the changepoint variable to state 3. Rather, the potentials corresponding to such evidence must be multiplied onto the messages as prescribed by the belief propagation algorithm (see Equation 22).

**Non-stationarity** Stationarity of conditional probability distributions was used to simplify the formulae in the previous exposition, but is not required. All that is needed is to keep running products of respective probabilities instead of the powers in the exponents of $P_k, \widehat{P}_k$. They need to be computed incrementally and tabulated to avoid hidden linear terms in the computation.

**Extension to trees** The extension of PCC-DBN to causal trees (PCT-DBNs) is now fairly straightforward. Because each node $X_k$ can now have multiple children $Ch(X_k)$, we must replace $\Sigma_k$ in all recurrences with the true $\lambda$-potential for $X_k$:

$$\lambda_{X_k} = \lambda_{X_k \to X_k} \cdot \prod_{i \in Ch(X_k)} \Sigma_i, \quad (22)$$

where $\lambda_{X_k \to X_k}$ accounts for evidence observed in $X_k$'s temporal chain. The vector $\lambda_{X_k \to X_k}$ is zero where the evidence rules out a changepoint — before the time $t$ of the last observed $X_k^t = 0$ and after the time $s$ of first observed $X_k^s = 1$. Everywhere else, $\lambda_{X_k \to X_k}(j_k) = 1$. Note that $\lambda_X(x)$ potential can be obtained in $O(M)$ time per node.

In computation of $\psi$ potentials, $\Psi_k$ on the right-hand side of the recurrences is replaced by the $\pi_{Pa(X_k) \to X_k}$, which in turn include the influence of evidence under $X_k$'s siblings:

$$\pi_{Pa(X_k) \to X_k} = \pi_{X_k} \cdot \prod_{i \in Ch(Pa(X_k)) \setminus X_k} \Sigma_i. \quad (23)$$

Again, this preserves the $O(M)$ per-node complexity. Thus, PCT-DBN is linear in both $N$ and $M$.

## 4 Further Generalizations

In this section we describe three more important generalizations of PC-DBNs: polytrees, non-persistent nodes, and finally an approximate algorithm for general DAGs. These relaxations are more involved than those of Section 3.5 and thus require more elaboration.

### 4.1 Polytrees

Belief propagation [16, 17] is a powerful framework for exact inference in polytree networks. Polytrees, unlike trees, allow multiple parents of a node, but remain acyclic in the undirected sense. In polytree changepoint networks, structure in the conditional probability table $P(X = x|U)$ can be exploited to save a multiplicative factor of $M + 1$ just as we showed for tree networks. The $\psi$-recurrences run over the first parent variable, while the remaining parents are summed over by brute force. Similarly, the $\sigma$-recurrences run over the parent that the message is addressed to. For instance, the definition of $\sigma$ will be replaced by

$$\sigma_k^i \propto \sum_{L \leq j_k < M} \left[ \prod_{z=1}^{L} \widehat{P}_{\overline{k}}^{(z)} \right] \cdot \left[ \prod_{z=L+1}^{j_k} P_{\overline{k}}^{(z)} \right] P_k^{(j_k+1)} \cdot \Sigma_{k+1}^{j_k} \quad (24)$$

and Equation 11 by

$$\sigma_k^{i-1} = \sigma_k^i \cdot \frac{P_{\overline{k}}^{(i)}}{\widehat{P}_{\overline{k}}^{(i)}} + \left[ \prod_{z=1}^{i-1} \widehat{P}_{\overline{k}}^{(z)} \right] \cdot P_k^{(i)} \cdot \Sigma_{k+1}^{i-1}$$

$$\sigma_k^M = 0, \quad (25)$$

where, assuming we are sending to the first parent,

$$P_k^{(z)} = P(X_k = 1 | U_1 = 1, I\{\mathbf{U}_{2:} \leq z\})$$
$$\widehat{P}_k^{(z)} = P(X_k = 1 | U_1 = 0, I\{\mathbf{U}_{2:} \leq z\}) \quad (26)$$

are now functions of the joint configuration of the remaining parents $\mathbf{U}_{2:}$. The proportionality constant in Equation 24 equals the product of the remaining parents' $\pi$-messages. We call this PPT-DBN, the persistent polytree algorithm.

The worst-case time complexity of PPT-DBN is dominated by the cost associated with the largest family-clique: $O((M+1)^{U_{max}})$. The $2TBN$ algorithm [12] suffers a worst-case time complexity $O(2^{2N}M)$, as all nodes in two slices may be entangled [9] in the clique to connect the two subsequent time-slices, even though the prototype network is a polytree [12, section 3.6.2]. Therefore, we expect PPT-DBN will be comparatively better for shorter temporal chains of larger networks. However, PPT-DBN really shines on space complexity. At most $O(MN)$ memory is consumed, compared to 2TBN, where the potentials in the joint tree can grow as large as $O(2^{2N})$. Later we show experimentally how dramatic the difference can be.

### 4.2 Non-persistent nodes

While it is convenient to assume that all non-leaf variables are persistent, it does limit the modeling power at our disposal. We now show how an occasional non-persistent variable in the network can also be handled in polynomial time. We assume the non-persistent variable is isolated, that is, all of its neighbors are persistent. We make this assumption in order to avoid having to invoke an embedded general DBN inference

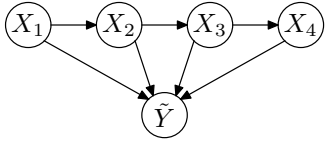

Figure 2: The minimal example of network with a non-persistent node.

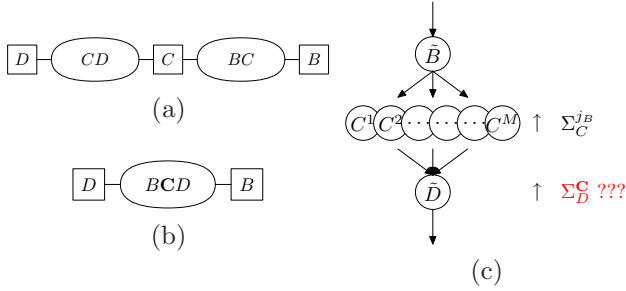

Figure 3: **a)** Induced cliques and separators **b)** Enlarged clique for generalized BP **c)** Σ-message flow in a network with a non-persistent variable

algorithm such as 2TBN to handle connected non-persistent variables. It can be done, but quickly becomes complex and inelegant. A simple way to handle connected non-persistent nodes is to combine them into a single joint node. Obviously, this solution causes exponential growth in the state space of the joined nodes, making it somewhat unappealing. A two-slice approach made aware of the determinism, e.g. by use of ADD compilation [3, 2], could very well work better for networks with only a few persistent variables.

### 4.2.1 A simple example

To illustrate how an isolated non-persistent node would be handled, assume first a simple structure such as in Figure 2. Then we can efficiently compute $P(\tilde{Y} = j)$ by moving the sums inward:

$$P(\tilde{Y} = j) = \sum_{\mathbf{X}} P(\mathbf{X}) P(\tilde{Y} = j | \mathbf{X}) =$$
$$\sum_{X^1,...X^M} \left[\prod_{k=1}^{M} P(X^k | X^{k-1})\right] \left[\prod_{k=1}^{M} \phi_j^k\right] =$$
$$\underbrace{\sum_{X^1} P(X^1) \phi_j^1 \underbrace{\sum_{X^2} P(X^2 | X^1) \phi_j^2 \ldots \underbrace{\sum_{X^M} P(X^M | X^{M-1}) \phi_j^M}_{\kappa^M}}_{\kappa^2}}$$

This gives rise to the recurrence

$$\kappa_i^M(j) = \sum_v P(X^M = v | X^{M-1} = i) \cdot \phi_j^M \quad (27)$$
$$\kappa_i^k(j) = \sum_v P(X^k = v | X^{k-1} = i) \cdot \phi_j^k \cdot \kappa_v^{k+1},$$

with $\phi_j^k$ defined appropriately:

$$\phi_j^k(X^k) = \begin{cases} P(Y^1 = 0 | X^1) & \text{if } k = 1 \\ P(Y^k = 0 | Y^{k-1} = 0, X^k) & \text{if } 1 < k \leq j \\ P(Y^k = 1 | Y^{k-1} = 0, X^k) & \text{if } k = j + 1 \\ 1 & \text{if } k > j + 1 \end{cases}$$

Now, $P(\tilde{Y} = j) = \kappa_1^0$. Moreover, a short analysis will reveal that

$$\kappa_i^k(j) = \kappa_i^{k+1}(j+1) \qquad \forall 1 \leq k < j < M - 1.$$

Therefore, we do not need to compute $\kappa$ for every $j$, but compute $\kappa_i^k(M)$ for all $k$ as a special case and then $\kappa_i^k(M-1)$ for all $k$ to start the recursion. All other values can be read off $\kappa_i^k(M-1)$ with the appropriate indexing shift. Thus, we can obtain the entire distribution $P(\tilde{Y})$ in $O(M)$ time! Allowing non-persistent variables to take on multiple values is also straightforward: we only need to allow the bottom index in $\kappa_i^k$ to range over the domain of $X_k$.

### 4.2.2 The general case

Pearl's belief propagation has been generalized to the clique tree propagation algorithm [21]. With belief propagation (BP), the cliques correspond to edges of the original polytree and the separators consist of single nodes. In the process of message passing, the variables not in the separator are summed out of the clique potentials.

The PCT-DBN algorithm used the "natural" cliques induced by the transformed network. Assume we have a situation such as in Figure 3. Because the variable $C$ is not persistent, the size of the induced separator $\boxed{\mathbf{C}}$ is $2^M$. However, we can work conceptually with a larger clique $B\mathbf{C}D$. Message propagation then calls for summing out all $C^{1:M}$, which we can do without actually instantiating the clique potential using a recurrence derived much like that of Equation 27. In the interest of space, we only show here the simple $\kappa$ recurrence. The full recurrence calls for summing over all persistent variables in the clique and the resulting complexity is $O((M+1)^B)$, where $B$ is the number of persistent neighbors of the non-persistent variable.

## 5 Experimental evaluation

We implemented our algorithms in Matlab and compare them to the exact and approximate algorithms as implemented in the Bayesian Network Toolbox (BNT) [11]. Namely, we will compare to the Boyen-Koller (BK) algorithm [1] in its 1) exact and 2) fully factored setting. Although BK reduces in its exact form to the incremental junction-tree algorithm, we found it was faster in practice than the 2TBN implementation.

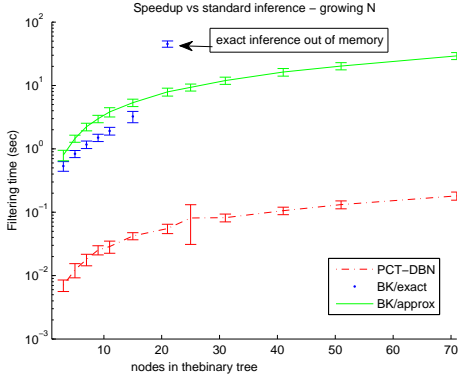

Figure 4: Performance scale up of PCT-DBN with $N$. The temporal length was fixed at $M = 20$. Note log scale $y$-axis.

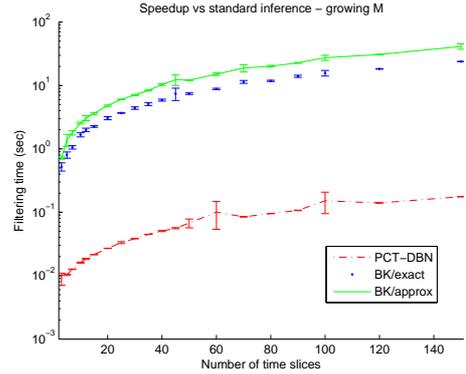

Figure 5: Performance scale up of PCT-DBN with $M$. The number of nodes was held at $N = 19$.

Therefore the 2TBN algorithm is not included in the evaluation.

Matlab run-time is not the ideal measure of algorithm complexity as it is arguably more sensititve to the quality of implementation compared to other languages. However, we should note that we did not make any special effort to optimize our code for Matlab, and the BNT library is a widely used and mature code base, so we expect any advantages due to code quality to fall to the competing approaches. Our Matlab code and further evaluation results can be downloaded at http://www.cs.pitt.edu/~tomas/papers/UAI08.

### 5.1 Speed of the tree algorithm

To compare inference speed, a network with the structure of a full binary tree with $N$ nodes was generated. Among the $MN$ possible observations, 10% of the variables were set to a random value (subject to persistence constraints so that $P(E) \neq 0$). We measured the time to execute the query $p(\tilde{X}_1|E)$—the posterior probability over the root node—for each algorithm[2]. This process was repeated 100 times for each $M$, $N$ combination and the respective times added up. The results are graphed out in Figures 4 and 5.

PCT-DBN outperforms both the exact incremental joint tree algorithm and the approximate BK algorithm (assuming independence) by several orders of magnitude as $N$, the size of a slice grows (Figure 4). In fact, the exact algorithm soon runs out of memory (around $N = 20$) and only the approximate version keeps up. Exact PCT-DBN inference also performs consistently about 100 times faster than exact junction-tree and approximate BK inference when we look at scale-up with the number of slices, as shown in

---
[2]The actual query is in fact irrelevant as all algorithms compute all posterior marginals simultaneously.

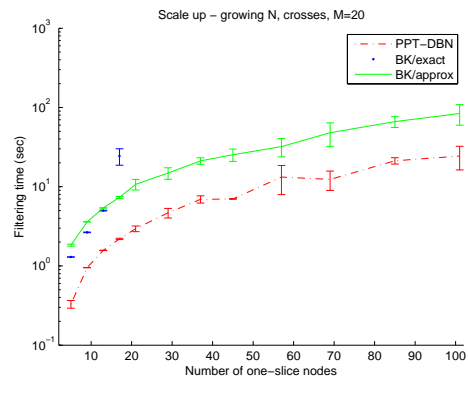

Figure 6: Performance scale up of PPT-DBN with $N$.

Figure 5.

### 5.2 Speed of the polytree algorithm

The asymptotic time complexity of PPT-DBN, as $M$ increases, may be less favorable than that of the incremental approaches. However, its lower memory complexity is very favorable, as documented by the following experiment. We generated a network where most non-root nodes have exactly 2 parents and measured the time for the three inference algorithms. Quadratic scale-up with $M$ is expected for PPT-DBN in such a network.

Figure 6 shows the *exact* PPT-DBN algorithm to be several times faster than, but scaling very similarly to, the *approximate* fully factorized Boyen-Koller algorithm with an $M = 20$ time slice inference window. Peeking ahead into Figure 7 suggests the time performance would be about identical at $M = 70$ time slices. The junction-tree algorithm does not scale beyond 20 nodes due to memory usage.

Figure 7 shows clearly that asymptotically, PPT-DBN

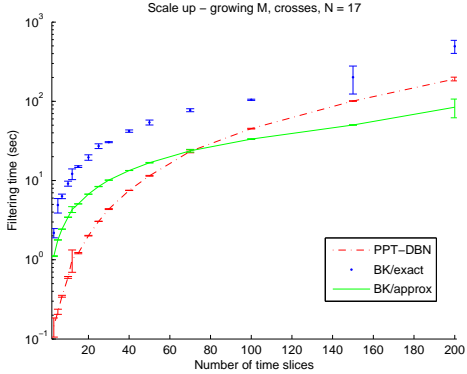

Figure 7: Performance scale up of PPT-DBN with $M$. The number of nodes was held at $N = 17$.

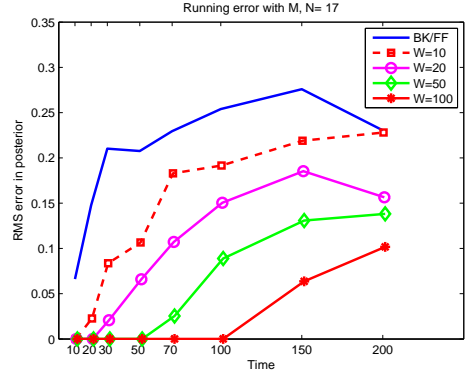

Figure 8: Accuracy of PC-DBN with growing inference window $M$. Averages are over 100 different parameterizations of the network; error bars are omitted for clarity but standard deviations are the same order of magnitude as the means for all curves.

scales with a steeper slope than both BK inference and junction-tree inference. Indeed, for about $N = 70$, BK eventually surpasses PPT-DBN in terms of speed. However, it remains faster than junction-tree incremental inference throughout the range. On a computer with 1 GB RAM, the exact version begins to hit memory limits around $M = 200$ and $N = 17$.

We conclude that if exact inference is desired for persistent polytree causal networks, using the PPT-DBN algorithm is a better choice for a wide range of inference window lengths. Furthermore, if approximate inference is acceptable, we show in Section 5.3 that for large enough $N$, fixed window smoothing using PPT-DBN can outperform BK inference in terms of RMS error, while still performing many times faster. For the special case of persistent causal trees, the new algorithm dominates by orders of magnitude in all ranges that we tested versus both junction tree and BK assuming intra-slice independence.

### 5.3 Fixed-window approximation

A minor disadvantage of the PPT-DBN algorithm is that it cannot do online inference yet. Therefore, when monitoring a process, $M$ grows and so does the computation time. In practice, only a fixed number of most recent observations are usually considered with older observations falling out of the "window". Thus we evaluate if reasonable precision can be attained with small window sizes, where PPT-DBN dominates.

Figure 8 shows, for several time slices $t$, the root mean square error of computed posterior marginals

$$Err^t_{BK} = \sqrt{\sum_{i=1}^{N}[P_{BK}(X_i^t|O^{1:t}) - P_{ex}(X_i^t|O^{1:t})]^2}$$

incurred by the fully factored Boyen-Koller method and the same error for PPT-DBN which ignores all evidence older than $W$, for different values of $W$. We use a binary tree prototype with all leaf variables non-persistent and observed. All non-leaf variables are persistent and hidden. The CPT probabilities are sampled uniformly at random. The observed evidence $O$ is obtained by forward-sampling the DBN and restricting it to the observables. We find that the error of our algorithm falls with growing $W$ as expected. The results become even more favorable for PC-DBN as N, the number of nodes per slice, grows (see also further results online). The error made by fixing the inference window tends to be lower than that of the Boyen-Koller approximation for reasonable values of $W$ and we can eliminate the unfavorable dependence on $M$ at a small price of accuracy. One clear drawback to a naive implementation of the fixed-window approach is that if evidence is not observed at each time slice, in the presence of persistence a piece of crucial evidence might drop off the window preventing the model from "remembering" that a persistent state was already acheived. This glitch could in principle be fixed by caching when persistent variables have turned on.

## 6 Conclusions and future work

We presented an algorithm for PC-DBNs, a way to exploit the special structure of the DBN probability distribution when many variables are persistent. Unlike forward-backward approaches to DBN inference that work slice-to-slice, we collapse the entire temporal progression and perform inference in the original prototype network structure. For trees, the algorithm is many times faster than state-of-the-art general-purpose exact *and approximate* DBN inference algorithms, while having a space complexity of only

$O(MN)$. This continues to hold even in the polytree generalization with inference window lengths into the hundreds. While this method does not directly yield an incremental filtering algorithm, we show that a fixed-window smoothing version of PC-DBN inference can perform approximate filtering faster and with comparable or less error than BK-filtering.

Although we have not presented a filtering algorithm that can exploit persistence, we do believe that one is possible. The number of possible joint configurations of variables in two subsequent slices is $3^N$ with the persistence assumption as opposed to $4^N$ in the general network. This hints at the possibility of a 2TBN-like algorithm leveraging persistence and still remaining linear in the number of time slices.

Another possible direction for this work is to allow multi resolution temporal modeling by modeling systems on very short time scales, but utilizing a persistence approximation for the slow processes. In such cases, a model with a single time-scale could efficiently and accurately deal with systems that have both fast and slow processes.

Also interesting is the vision of approximate inference algorithms not requiring persistence, but simply assuming that the hidden state changes at most once in the period of interest. If the change in the hidden state is relatively slow, this could be a fairly accurate approximation. Such problems are often found in bioinformatics areas such as phylogeny discovery, where time of a mutation is of interest [6].